\documentclass[10pt,twocolumn,letterpaper]{article}

\usepackage[pagenumbers]{cvpr} 

\usepackage{adjustbox}
\newcommand{\cmark}{\textcolor{green!60!black}{\ding{51}}}
\newcommand{\xmark}{\textcolor{red!70!black}{\ding{55}}}
\usepackage{pifont}
\usepackage[table]{xcolor}
\definecolor{rowhighlight}{HTML}{D9E2F3} 
\usepackage{dblfloatfix}
\usepackage{caption}
\usepackage{placeins}        
\usepackage{float}

\definecolor{cvprblue}{rgb}{0.21,0.49,0.74}
\usepackage[pagebackref,breaklinks,colorlinks,allcolors=cvprblue]{hyperref}

\def\paperID{20587} 
\def\confName{CVPR}
\def\confYear{2026}

\title{InfinityStory: Unlimited Video Generation with World Consistency and Character-Aware Shot Transitions}

\author{
  \fontsize{8}{9}\selectfont
  \textbf{Mohamed Elmoghany}$^{1}$,
  \textbf{Liangbing Zhao}$^{2}$,
  \textbf{Xiaoqian Shen}$^{2}$,
  \textbf{Subhojyoti Mukherjee}$^{1}$,
  \textbf{Yang Zhou}$^{1}$,
  \textbf{Gang Wu}$^{1}$,\\
  \fontsize{8}{9}\selectfont
  \textbf{Viet Dac Lai}$^{1}$,
  \textbf{Seunghyun Yoon}$^{1}$,
  \textbf{Ryan Rossi}$^{1}$,
  \textbf{Abdullah Rashwan}$^{1}$,
  \textbf{Puneet Mathur}$^{1}$,
  \textbf{Varun Manjunatha}$^{1}$,\\
  \fontsize{8}{9}\selectfont
  \textbf{Daksh Dangi}$^{3}$,
  \textbf{Chien Nguyen}$^{1,4}$,
  \textbf{Nedim Lipka}$^{1}$,
  \textbf{Trung Bui}$^{1}$,
  \textbf{Krishna Kumar Singh}$^{1}$,
  \textbf{Ruiyi Zhang}$^{1}$,\\
  \fontsize{8}{9}\selectfont
  \textbf{Xiaolei Huang}$^{5}$,
  \textbf{Jaemin Cho}$^{6}$,
  \textbf{Yu Wang}$^{4}$,
  \textbf{Namyong Park}$^{7}$,
  \textbf{Zhengzhong Tu}$^{8}$,
  \textbf{Hongjie Chen}$^{9}$,\\
  \fontsize{8}{9}\selectfont
  \textbf{Hoda Eldardiry}$^{10}$,
  \textbf{Nesreen Ahmed}$^{11}$,
  \textbf{Thien Nguyen}$^{4}$,
  \textbf{Dinesh Manocha}$^{12}$,\\
  \fontsize{8}{9}\selectfont
  \textbf{Mohamed Elhoseiny}$^{2}$,
  \textbf{Franck Dernoncourt}$^{1}$\\[1ex]
  \fontsize{8}{9}\selectfont
  $^{1}$Adobe Research \quad
  $^{2}$KAUST \quad
  $^{3}$Independent Researcher \quad
  $^{4}$University of Oregon\\
  \fontsize{8}{9}\selectfont
  $^{5}$University of Memphis \quad
  $^{6}$Johns Hopkins University \quad
  $^{7}$Meta AI \quad
  $^{8}$Texas A\&M University\\
  \fontsize{8}{9}\selectfont
  $^{9}$Dolby Labs \quad
  $^{10}$Virginia Tech \quad
  $^{11}$Cisco \quad
  $^{12}$University of Maryland, College Park
}

\newcommand{\model}{InfinityStory\xspace}

\begin{document}

\twocolumn[{
\maketitle
\begin{center}
\includegraphics[width=0.95\textwidth]{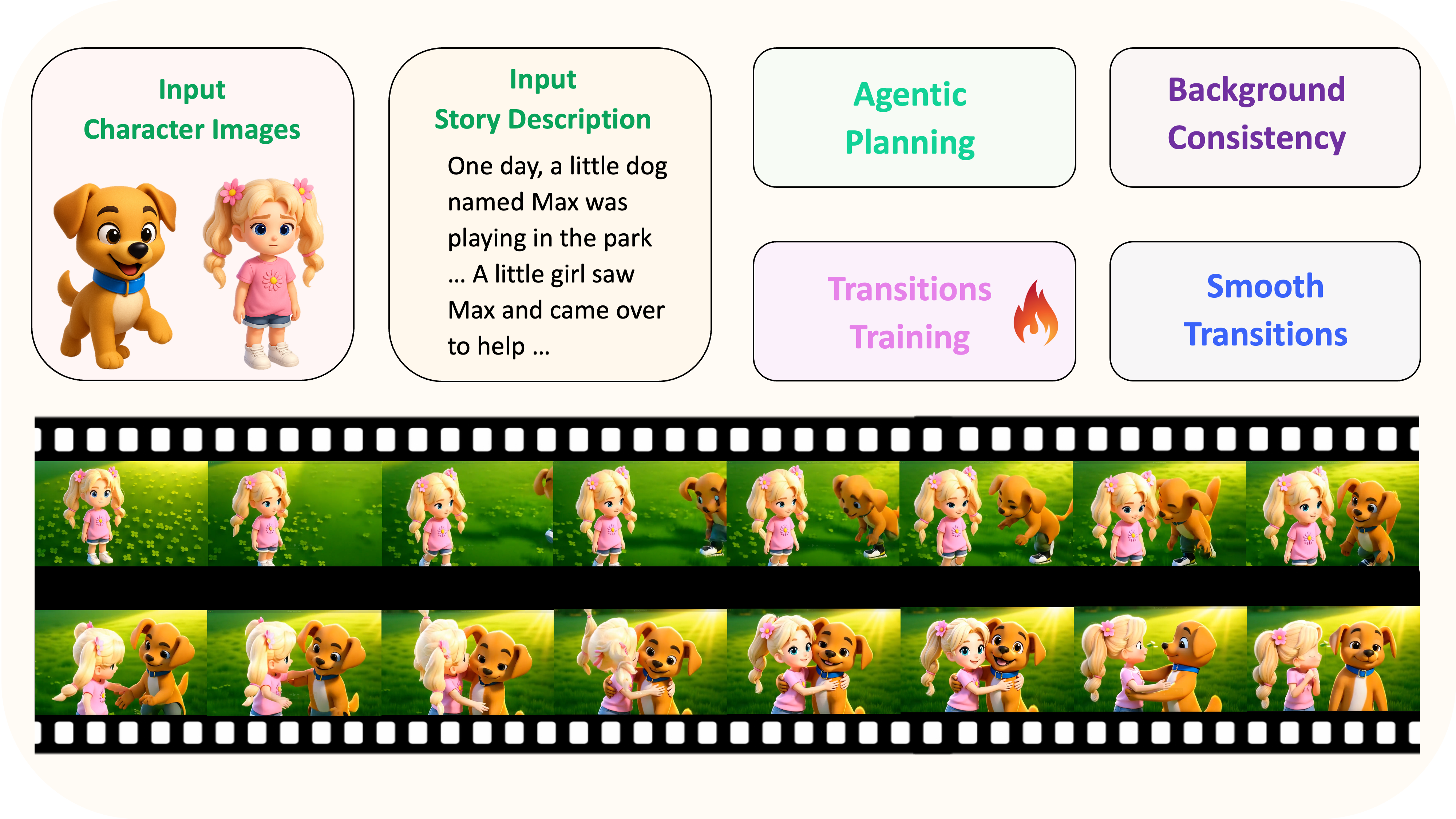}
\captionof{figure}{
Our pipeline takes an input of story text along with characters' reference images. It outputs a long video with background consistency and smooth transitions.
The figure shows our pipeline capability of generating multi-shot scene with consistent background.
It also shows a smooth transition between different shots in one scene (smooth shot-to-shot transitions) emphasizing that multi-subject characters do not appear or disappear suddenly. Our model is the first of its kind in multi-subject smooth transitions.
}
\label{fig:teaser}\label{fig:teaser}
\end{center}
}]    

\begin{abstract}
Generating long-form storytelling videos with consistent visual narratives remains a significant challenge in video synthesis. We present a novel framework, dataset, and a model that address three critical limitations: background consistency across shots, seamless multi-subject shot-to-shot transitions, and scalability to hour-long narratives. Our approach introduces a background-consistent generation pipeline that maintains visual coherence across scenes while preserving character identity and spatial relationships. We further propose a transition-aware video synthesis module that generates smooth shot transitions for complex scenarios involving multiple subjects entering or exiting frames, going beyond the single-subject limitations of prior work. To support this, we contribute with a synthetic dataset of 10,000 multi-subject transition sequences covering underrepresented dynamic scene compositions. On VBench, InfinityStory achieves the highest Background Consistency (88.94), highest Subject Consistency (82.11), and the best overall average rank (2.80), showing improved stability, smoother transitions, and better temporal coherence. 

\begin{table*}[t]
\centering
\caption{Comparison of storytelling video generation frameworks.
}
\vspace{2mm}
\begin{adjustbox}{max width=\textwidth}
\begin{tabular}{lccccccc}
\toprule
\textbf{Method} & \textbf{Agentic Story} & \textbf{Shot Designing} & \textbf{ID Consistency} & \textbf{Background Consistency} & \textbf{Shot-to-Shot Transitions} \\
\midrule
LCT~\cite{longcontexttuning}                 & \xmark & \xmark & \cmark & \xmark & \xmark \\
Captain Cinema~\cite{captaincinema}          & \cmark & \cmark & \cmark & \xmark & \xmark \\
MovieAgent~\cite{movieagent}                 & \cmark & \cmark & \cmark & \xmark & \xmark \\
MAViS~\cite{mavis}                           & \cmark & \cmark & \cmark & \xmark & \xmark \\
\rowcolor{rowhighlight}
\textbf{InfinityStory (Ours)}                & \cmark & \cmark & \cmark & \cmark & \cmark \\
\bottomrule
\end{tabular}
\end{adjustbox}
\end{table*}
\end{abstract}    

\section{Introduction}
\setlength{\abovecaptionskip}{4pt}
\begin{figure*}[t]
\centering
\includegraphics[width=\textwidth]{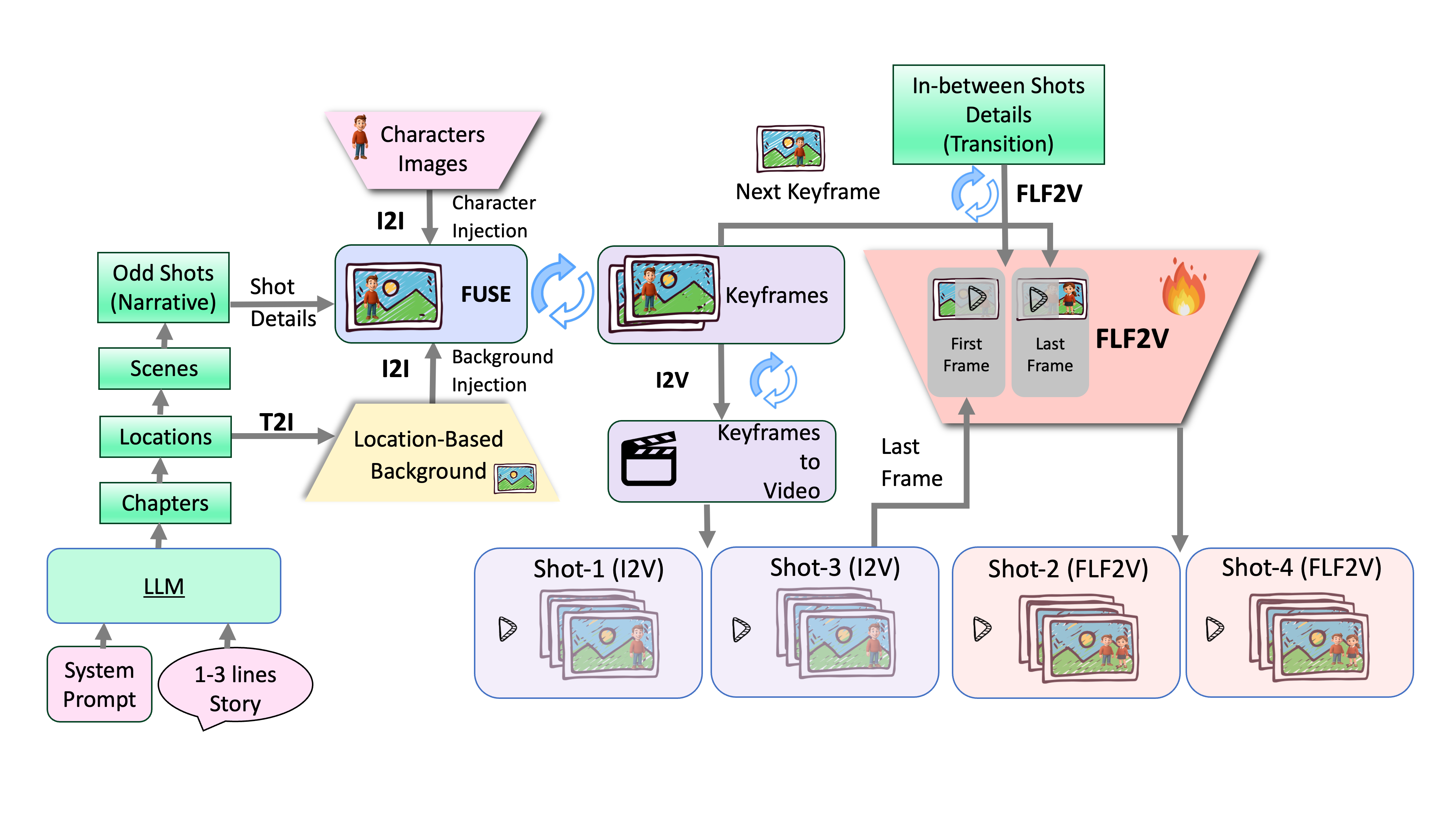}
\caption{Overview of the proposed storytelling video generation pipeline. \textcolor{green!60!black}{Green shapes:} are the output of the agentic pipeline. \textcolor{magenta!60!black}{Purple Shapes:} Narrative odd shots generate keyframe images which are used to generate video shots using I2V. \textcolor{red!60}{Red shapes:} While the transition in-between (even) shots take the next keyframe and the last frame from the generated I2V shot to generate a First-Last-Frame-to-Video (FLF2V) which smoothly bridges consecutive narrative shots. The output video would be stitched together to form one coherent video, i.e., shot-1 (I2V) $\to$ shot-2 (FLF2V) $\to$ shot-3 (I2V) $\to$ shot-4 (FLF2V) $\to$ .. and so on.
}
\label{fig:pipeline}
\end{figure*}
\setlength{\abovecaptionskip}{10pt}
Storytelling through images and videos plays a central role in entertainment, education, and digital media. Visual stories require consistent characters, coherent scenes, and clear narrative progression across many frames. Recent work such as StoryGPT-V~\cite{storygptv} shows that multimodal large language models can generate hundreds of coherent image frames for long-form story visualization. However, its outputs are discrete images rather than continuous video, it does not address cinematographic transitions, and it must be trained on a single movie domain. The creation of long-form storytelling videos remains a frontier in generative AI. Recent advances in text-to-video and image-to-video generation, specifically with open-source models such as CogVideo~\cite{cogvideo}, Wan~\cite{wan}, HunyuanVideo~\cite{hunyuanvideo}, and LTX-Video~\cite{ltxvideo}, have made videos more realistic and opened doors for cinematic long-form video research.

To manage the complexity of coherent video narratives, recent work adopts agentic multi-agent frameworks. MAViS~\cite{mavis} coordinates agents for script writing, shot design, and animation. MovieAgent~\cite{movieagent} uses hierarchical Chain-of-Thought planning with director, screenwriter, and cinematographer agents. AniMaker~\cite{animaker} applies Monte Carlo Tree Search for efficient clip selection. MovieDreamer~\cite{moviedreamer}, Captain Cinema~\cite{captaincinema}, and FilMaster~\cite{filmaster} also employ hierarchical planning with keyframe generation and RAG-based cinematography.

These multi-agent approaches improve narrative structure and coordination, but they remain limited to short clips, typically 4–16 seconds, with no long-term visual consistency~\cite{mavis, movieagent, animaker}. Although agentic planning simplifies workflow, current generators lack explicit background preservation~\cite{v3gan, videostudio} and multi-subject transition modeling~\cite{cinetrans, cut2next}, causing scene drift and jarring boundaries when clips are concatenated.

\noindent\textbf{Background Consistency Challenge.} 
background preservation remains an underexplored aspect of long-form video generation. Methods such as V3GAN~\cite{v3gan}, VideoStudio~\cite{videostudio}, and ConsistI2V~\cite{consisti2v} focus on foreground character consistency while treating backgrounds implicitly. When backgrounds are regenerated together with moving subjects, small deviations accumulate across shots, leading to drift in appearance, lighting, and spatial layout. This becomes problematic in long narratives where scene identity must remain stable over hundreds of shots. Approaches like MovieAgent~\cite{movieagent}, Long Context Tuning~\cite{longcontexttuning}, AnimeShooter~\cite{animeshooter}, and Captain Cinema~\cite{captaincinema} rely on prompting the diffusion model with background descriptions, which causes the background and location to change from shot to shot, breaking visual coherence.

\noindent\textbf{Shot-to-Shot Transition Challenge.} Beyond background stability, smooth cinematographic transitions between shots remain largely unexplored. Most pipelines generate independent clips and concatenate them, producing jarring edits. While CineTrans~\cite{cinetrans} and HoloCine~\cite{holocine} advance transition quality through attention-guided diffusion and structured camera modeling, they lack multi-subject transitions, where multiple characters enter, exit, or replace each other between shots as characters suddenly appear or disappear with a new shot. Underrepresented cases lacking synthesis mechanisms for coordinating character movements, spatial relationships, and background continuity.

To address these challenges, we introduce \textbf{InfinityStory}, a framework for scalable, cinematic, multi-shot video generation.
Our framework is built on two key ideas.
\textit{Background consistency.} We generate a fixed set of story locations and assign each scene to one location. For every shot, we fuse the characters with the corresponding background to preserve scene identity. This ensures stable backgrounds across long sequences. Figure~\ref{fig:teaser} shows examples with consistent locations and multi-subject interactions.
\textit{Shot-to-shot transitions.} We build a dataset of 10,000 multi-subject transition videos covering entry, exit, and replacement events, filtered to avoid sudden character appearance or disappearance. A first–last-frame-to-video transition model trained on this dataset provides smooth, cinematographic transitions between shots.

InfinityStory achieves the strongest overall performance on VBench, ranking first with an average rank of 2.80, and obtains the highest Subject Consistency (82.11) and Background Consistency (88.94) among all baselines.

\noindent\textbf{In summary, our key contributions are:}
\begin{itemize}
    \item We introduce Cinematic Multi-Subject Transition Synthesis (CMTS). CMTS is a new method for generating smooth and cinematic shot-to-shot transitions. It handles multi-subject events such as character entry, exit and replacement with movement-type supervision. We provide a transition model trained to generate high-quality transition clips, along with a synthetic dataset of 10,000 annotated videos. To the best of our knowledge, no prior work systematically addresses multi-subject cinematographic transitions with entry/exit/replacement supervision.

    \item We propose an end-to-end framework that scales narrative video generation to hour-long sequences while preserving story structure and cinematographic coherence across hundreds of shots.
    \item We develop a background-consistent generation pipeline that decouples background preservation from foreground dynamics. It uses persistent location injection and selective cross-shot memory gating to give user-controllable scene stability over long sequences.
    \item We conduct extensive evaluations of state-of-the-art baselines and ablations under background-consistency and multi-subject transition settings. Existing methods struggle with persistent backgrounds and clean shot-to-shot transitions. Our approach improves stability and transition smoothness on automatic metrics and human studies.

\end{itemize}

\setlength{\abovecaptionskip}{4pt}

\begin{figure*}[t]
\centering
\includegraphics[width=0.8\textwidth]{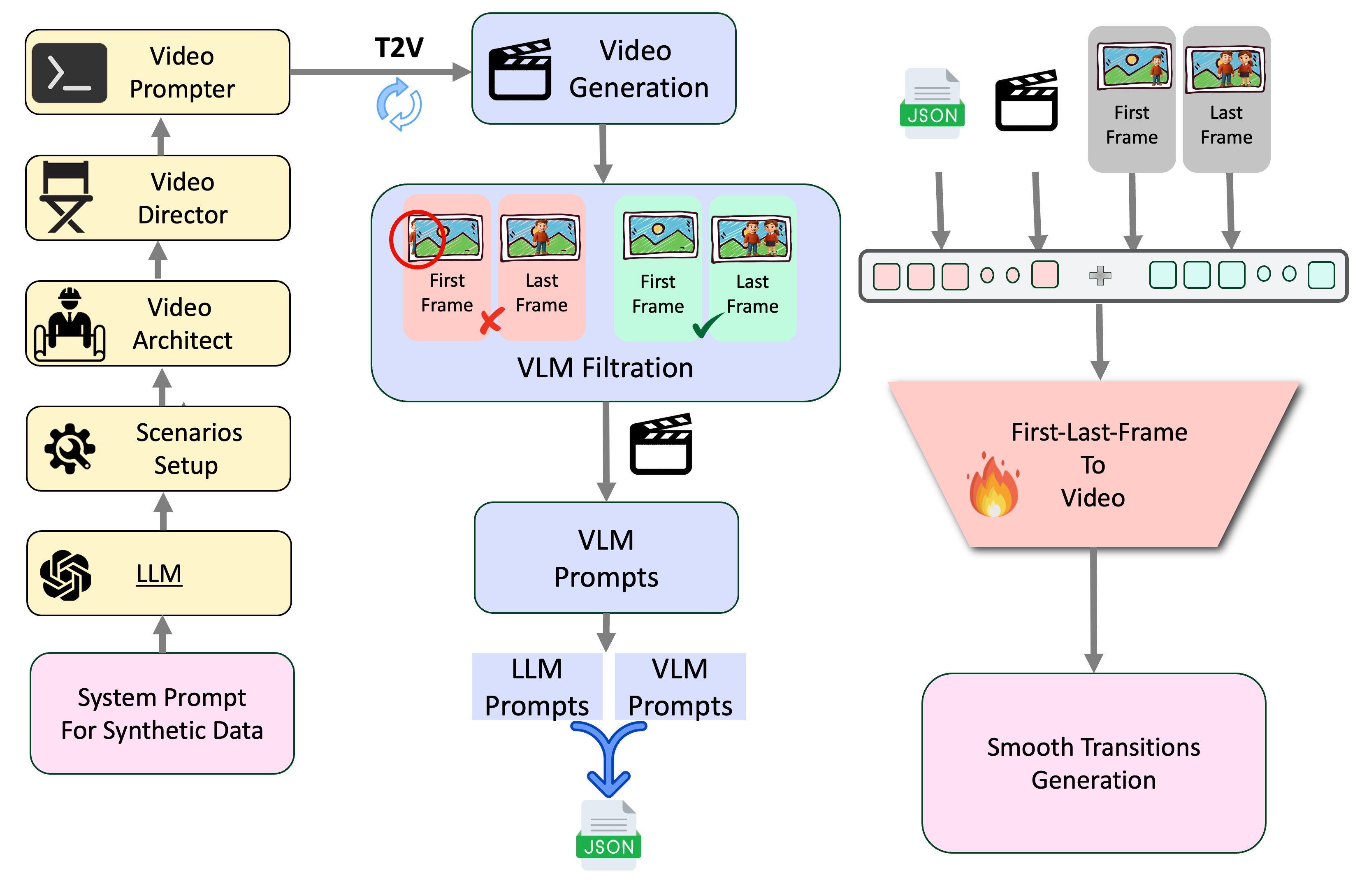}
\caption{This novel framework is designed to create a large dataset for smooth transitions to train the First-Last-Frame-to-Video (FLF2V). We use 4 agentic pipelines to generate the prompt that generates the transition video. Then we use VLM to filter out bad videos which does not have the correct number of characters and to generate another prompt for the transition video. We use the combination of prompts from the video prompter and VLM along with the generated video, first-frame and last-frame of the video to train the FLF2V.
}
\label{fig:transitions}
\end{figure*}
\setlength{\abovecaptionskip}{10pt}

\section{Related works}
\label{sec:related}

\noindent\textbf{Multi-Agent Planning.}
To handle long-form generation, recent works adopt hierarchical multi-agent pipelines. MovieAgent~\cite{movieagent} uses director, scene, and shot agents; MAViS~\cite{mavis} adds an initial script-writing stage; and AniMaker~\cite{animaker} generates multiple candidate clips per shot to select the best one. Unlike these methods, which rely on prompt consistency and camera-cut transitions, InfinityStory employs location-grounded planning and multi-subject transition modeling to maintain background consistency and produce smooth shot-to-shot transitions.

\noindent\textbf{Story Generation Background Consistency.}
Story visualization began predominantly in the image domain, where the goal is to generate a sequence of coherent images that follow a narrative. StoryGPT-V~\cite{storygptv} advances this direction by enabling multimodal LLMs to generate hundreds of consistent image frames. However, its outputs remain discrete images rather than continuous video, and although it maintains background consistency, it must be trained and applied within a single-movie domain.

Moving from images to video, background and scene consistency require dedicated modeling. Long Context Tuning~\cite{long-context-tuning} uses long-context attention across shots with interleaved positional embeddings to co-model scenes. Captain Cinema~\cite{captaincinema} applies multi-shot interleaved conditioning for cross-shot coherence. CineTrans~\cite{cinetrans} adopts mask-based diffusion for cinematic transitions.
However, these video-based methods still struggle to preserve stable backgrounds, and the background often shifts after a few seconds due to the underlying generative model behavior.

In InfinityStory, we introduce a dedicated agentic stage for location-grounded background generation. We synthesize a fixed set of locations where the story takes place. We enforce that each scene occurs in a specific location. Multiple shots form a scene with a shared background layout. For each shot, we fuse the selected background with the characters to construct a consistent keyframe image, which anchors the rest of the video generation pipeline.

\noindent\textbf{Shot-to-Shot Transitions.}
A critical yet underexplored aspect is synthesizing natural shot transitions following cinematographic conventions. The definition of Shot-to-shot transitions here is the handling of abrupt appearance or disappearance of characters in two consecutive shots. No characters are allowed to appear or disappear suddenly except when moving from one scene to another scene. CineTrans~\cite{cinetrans} Explicitly models cinematic transitions (cuts, pans, fades) via mask-based diffusion. Video-Gen-of-Though~\cite{videogenofthought} uses cross-shot latent propagation and adjacent latent transition modules manage smooth evolution between shots. Long Context Tuning~\cite{long-context-tuning} uses long-context attention across shots allows smooth frame-to-frame context blending but mainly uses camera to abruptly zoom in, zoom out. MovieAgent~\cite{movieagent} pipeline enforces logical and visual continuity at shot boundaries via LLM agents. While these methods enable transitions, the generated videos with near cosine similarity, they still have characters appearing or disappearing suddenly every 5 seconds breaking the shot-to-shot transitions rules. However, our work InfinityStroy, we generate a multi-subject character dataset of unrepresented data to have a smooth transitions between shots. 
\section{Methods}
\label{sec:methods}

Our framework enables scalable narrative video generation spanning hour-long sequences through a multi-agent system (Sec.~\ref{sec:problem},~\ref{sec:multiagent}) that addresses the critical challenges: \textbf{(i)} Enhancement of the background consistency across all shots within a scene via location-based background and character injection (Sec.~\ref{sec:background}), and \textbf{(ii)} Towards smooth multi-character shot-to-shot transition that prevent abrupt character appearances or disappearances between consecutive shots (Sec.~\ref{sec:transitions}). To train models capable of handling multi-subject transitions, we introduce a synthetic dataset targeting underrepresented transition scenarios (Sec.~\ref{sec:dataset}). Figure~\ref{fig:transitions} illustrates our approach.

\subsection{Problem Formulation}
\label{sec:problem}

Given a high-level story specification $c$, our multi-agent system (Sec.~\ref{sec:multiagent}) decomposes the generation task hierarchically: chapters $Ch=\{ch_1,\dots,ch_{N_{\text{ch}}}\}$ → locations $\Lambda=\{\ell_1,\dots,\ell_M\}$ → scenes $Sc=\{sc_1,\dots,sc_{N_{\text{sc}}}\}$ → shots $Sh=\{sh_1,\dots,sh_K\}$. Each scene binds to a fixed location from $\Lambda$, ensuring background consistency across all shots within that scene (Sec.~\ref{sec:background}). Each shot $k$ produces a video clip $V_k \in \mathbb{R}^{T_k\times H \times W \times 3}$ where $T_k$ denotes the number of frames. Odd-indexed shots use Image-to-Video (I2V) for narrative content; even-indexed shots use a finetuned First-Last-Frame-to-Video (FLF2V) for transitions with explicit transition metadata $\tau_k$. The transition metadata include information about how the character will enter/exit/stay in the shot in a narrative coherence (Sec.~\ref{sec:transitions}).

To enable smooth multi-character transitions, we generate a synthetic dataset of 10,000 transition videos targeting underrepresented scenarios such as character entry, exit, and multi-character replacements (Sec.~\ref{sec:dataset}). The FLF2V model is then finetuned on this dataset to ensure characters naturally appear, disappear, and move between frames without abrupt jumps or discontinuities (Sec.~\ref{sec:transitions}).

\subsection{Multi-Agent Narrative for Long-Video Planning}
\label{sec:multiagent}

Given a high-level story specification $c$ (short story description, character names, images and descriptions), our multi-agent system decomposes the generation task through a four-stage hierarchical pipeline with explicit constraint propagation. Each agent is implemented as a structured LLM~\cite{openai2025gpt5} that produces narrative plans with internal reasoning traces, enabling programmatic retrieval, validation, and constraint checking.

\noindent\textbf{Chapter Agent.} Analyzes narrative structure to partition the story into $N_{\text{ch}}$ chapters $Ch=\{ch_1,ch_2,\dots,ch_{N_{\text{ch}}}\}$ (typically 10-20), each representing a major plot beat or turning point. The agent outputs narrative structure containing : (1) Relationships across all character pairs  (2) creative planning notes documenting narrative arc decisions, and (3) per-chapter metadata including plot summary, involved characters, timeline annotation (for example "Day 1, Morning"), and division justification. These are the global constraints along with character relationships and the temporal ordering to propagate to downstream agents. 

\begin{figure*}[t]
\centering
\includegraphics[width=0.97\textwidth]{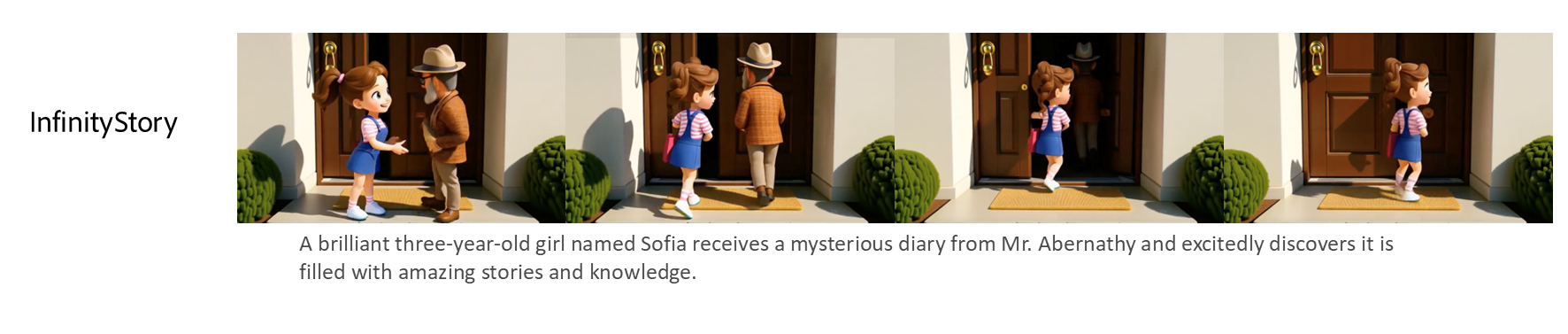}
\caption{Our pipeline results in smooth transitions}
\label{fig:qual-nmain}
\end{figure*}

\noindent\textbf{Location Agent.} Constructs a reusable location library $\Lambda=\{\ell_1,\dots,\ell_M\}$ by analyzing all chapters to identify recurring setting types. Each location $\ell$ receives a one-word name (e.g., "Castle", "Forest") and a character-free background description $b_{\ell}$ (2-3 sentences) covering architecture, lighting, weather, and mood. The agent documents location consolidation strategy to ensure sufficient variety across indoor/outdoor, day/night, and emotional atmospheres. Unlike prior work~\cite{movieagent} which generates shots without explicit binding, our pre-computed library enables: (1) consistent background reuse across all shots within a scene, and (2) programmatic anti-repetition constraints (consecutive scenes must use different locations), scaling to hour-long sequences without per-shot regeneration overhead.

\noindent\textbf{Scene Agent.} Expands each chapter $ch_i\!\in\!Ch$ into scenes $Sc=\{sc_1,\dots,sc_{N_{\text{sc}}}\}$ under strict constraints inherited from upstream agents. Each scene $sc_j$ must: (1) select exactly one location $\ell_j\!\in\!\Lambda$ matching the chapter's narrative needs, (2) use only characters from the chapter agent's allowed set $\mathcal{C}_j\!\subseteq\!\mathcal{C}$, (3) contain an odd shot count to enforce alternating I2V/FLF2V structure, and (4) use a different location than the adjacent scene. The agent outputs director's planning notes plus per-scene metadata: plot summary, exact location name and description copied from $\Lambda$, emotional tone, key props, music/sound cues, and cinematography notes. This location binding ensures spatial coherence: all shots in $sc_j$ share the same background $b_{\ell_j}$.

\noindent\textbf{Shot Agent.} Decomposes each scene into shots $Sh=\{sh_1,\dots,sh_K\}$ with alternating generation types: odd-indexed shots use I2V for narrative content, even-indexed shots use FLF2V for character transitions. The agent enforces a fixed vocabulary for programmatic validation: emotions $\in\{$Neutral, Angry, Happy, Sad$\}$, poses $\in\{$Standing, Sitting, Walking, Running, Reaching$\}$, interactions $\in\{$None, Shaking Hands, Hugging, Talking, Handing Over an Object$\}$, and at most 2 characters per shot. Each shot directive $sh_k$ contains: (1) character emotion, pose, interaction dictionaries, (2) dialogue with subtitle text and timing (0-5s), (3) plot visual description reusing the scene's background, (4) shot type and camera movement, (5) keyframe prompt for I2V to inject the characters into location-based background scene along with character emotion, and (6) video prompt describing 5-second action progression. Critically, even-indexed FLF2V shots include explicit transition metadata $\tau_k$: previous, current, next shot character sets, entering/exiting character lists and movement type $\in\{$Entry, Exit, No Change, Combination$\}$, and transition description detailing choreography for example, character walks in from outer frame toward center. This structured transition logic is absent in prior work~\cite{movieagent,mavis,filmaster}, enables smooth character entry/exit with explicit movement paths from or toward frame edges rather than abrupt appearance/disappearance.

\noindent\textbf{Sequential Generation with Memory.}   Given validated plans $\{sh_k\}$ and global context $c$, generation proceeds sequentially. Each shot conditions on normalized directive $\tilde{sh}_k=f(sh_k,c)$ and transition metadata $\tau_k$ (for FLF2V shots). Unlike prior work~\cite{movieagent,mavis,filmaster} which used camera sequences where characters abruptly appear and disappear, our approach handles smooth multi-subject transitions. The sequence likelihood factorizes via cross-shot memory $\mathcal{M}$ that accumulates identity and layout information~\cite{longcontexttuning,storydiffusion} and $V_k$ is the generated video clip given shot $k$:
\[
p(\{V_k\}\mid \{sh_k\},c)=\prod_{k=1}^{K} p\!\big(V_k \,\big|\, \tilde{sh}_k,\, \tau_k,\, \mathcal{M}_{k-1}\big),
\]
\[
\mathcal{M}_{k}=\Phi\!\big(\mathcal{M}_{k-1},\,V_k,\,\tilde{sh}_k\big),
\]
where $\Phi$ updates memory to preserve identity and layout consistency while limiting drift, initialized as $\mathcal{M}_0=\mathbf{0}$. Rendering proceeds via modular components. T2I generates backgrounds $B_\ell$ for each $\ell\!\in\!\Lambda$. I2I keyframe compositor fuses $B_{\ell(k)}$ with character references to produce $K_k$. I2V generates odd shots $V_k$. FLF2V generates even shots $V_k$ from $(F_{k-1}^{T_{k-1}}, K_{k+1})$ using transition metadata $\tau_k$, where $F_k^t$ denotes frame $t$ of video $V_k$. Cross-shot memory $\mathcal{M}_k$ accumulates via $\Phi(\cdot)$. Prompts and output examples from each agent can be found in the appendix.

\subsection{Enhancing Background Consistency via Location Injection}
\label{sec:background}
A story should have different locations where the story actions occur. Within a scene, we ground consistency in the location-based background which remains consistent across all shots. Given the pre-generated location set $\Lambda$ and descriptions $\{b_{\ell}\}_{\ell\in\Lambda}$, a T2I~\cite{wan} produces a canonical background image $B_{\ell}$ for each location, where $g(\cdot;\phi)$ is the T2I and I2I generator function with model parameters $\phi$:
\[
B_{\ell}=g_{\mathrm{bg}}\!\big(b_{\ell};\,\phi\big),\qquad \ell\in\Lambda,
\]

For each scene bound to $\ell(k)$, we first generate the canonical T2I background $B_{\ell(k)}$ from the background description $b_{\ell(k)}$. Then, for shot $k$, we compose the keyframe via an I2I model~\cite{omnigen2} that fuses $B_{\ell(k)}$ with character reference images $R_k$:

\[
K_k = g_{\mathrm{i2i}}\!\big(B_{\ell(k)},\, R_k;\, \psi\big),
\]
where $\psi$ denotes the I2I model parameters and $R_k$ contains reference images of characters appearing in shot $k$. This keyframe $K_k$ embeds both the fixed background and character appearance, ensuring that the subsequent I2V generation inherits the injected background. Background stability is enforced by penalizing perceptual drift from $B_{\ell(k)}$ across frames $t$:
\[
\mathcal{L}_{\text{bg}} = \sum_{t=1}^{T_k}\big\| \phi(V_k^{t})-\phi\!\big(B_{\ell(k)}\big)\big\|_2^2,
\]
where $\phi(\cdot)$ is a pretrained visual encoder function~\cite{zhang2018perceptual} for perceptual loss and $V_k^t$ denotes frame $t$ of video $V_k$. Odd-indexed (narrative) shots are generated with I2V from $K_k$,
\[
V_k \sim p_{\mathrm{i2v}}\!\big(V \,\big|\, K_k,\, \tilde{sh}_k,\, \mathcal{M}_{k-1};\,\Theta_{\mathrm{i2v}}\big),\qquad k\ \text{odd},
\]
with last frame $F_k^{T_k}=\mathrm{last}(V_k)$, inheriting the fused background. Even-indexed (transition) shots are generated with FLF2V using the pair bridging adjacent narrative shots,
\[
V_k \sim p_{\mathrm{flf2v}}\!\big(V \,\big|\, F_{k-1}^{T_{k-1}},\, K_{k+1},\, \tilde{sh}_k;\,\Theta_{\mathrm{flf2v}}\big),\qquad k\ \text{even},
\]
where $K_{k+1}$ is composed with the same $B_{\ell(k)}$, locking background at both endpoints and reducing scene drift across transitions. This realizes the schedule $K_1\!\xrightarrow{\mathrm{i2v}}\!V_1,\ (F_1^{T_1},K_3)\!\xrightarrow{\mathrm{flf2v}}\!V_2,\ K_3\!\xrightarrow{\mathrm{i2v}}\!V_3$, aligning with the specified transition logic. 

\subsection{Towards Smooth Multi-Character Transition}
\label{sec:transitions}
Off-the-shelf FLF2V models~\cite{wan} are trained on interpolating between two images with consistent character presence, but fail on underrepresented scenarios: empty-to-occupied (character entry), occupied-to-empty (character exit), and multi-character replacements. We address this by introducing \emph{Cinematic Multi-Subject Transition Synthesis (CMTS)}, a novel challenge requiring smooth, cinematographic shot-to-shot transitions with explicit entry/exit/reposition supervision. To the best of our knowledge, no existing work systematically addresses multi-subject, cinematographic transitions with movement-type annotations.

Transition shots (even-indexed shots) are FLF2V transitions with explicit transition logic:
\[
\tau_k=\big(\mathcal{C}_{k-1},\,\mathcal{C}_{k}^{\text{start}},\,\mathcal{C}_{k}^{\text{end}},\,\mathcal{X}_{k}^{\text{exit}},\,\mathcal{E}_{k}^{\text{entry}},\,m_k\big),
\]
where $\mathcal{C}_{k-1}$ are characters in the previous odd shot 
$\mathcal{C}_{k}^{\text{start}}$, $\mathcal{C}_{k}^{\text{end}}$ are character sets at the start/end of the transition shot, $\mathcal{X}_{k}^{\text{exit}}$ and $\mathcal{E}_{k}^{\text{entry}}$ denote exiting/entering sets, and $m_k \!\in\!\{\text{Entry},\text{Exit},\text{No Change},\text{Combination}\}$. Figure~\ref{fig:transitions} describes our smooth multi-character transition pipeline.

\begin{table*}[!htbp]
\centering
\caption{Main results on TinyStories~\cite{tinystories}, presenting scores for Image Quality (Img Quality), Subject Consistency (Subject Cons.), Background Consistency (Bg Cons.), Aesthetic Quality (Aesthetic), Motion Smoothness (Motion Smth.), and Average Rank (Rank Avg. - the average ranking position across all models). InfinityStory achieves the best ranking average.}
\label{tab:vbench_evaluation}
\small
\setlength{\tabcolsep}{1pt}
\begin{tabular}{@{}lccccc|c@{}} 
\toprule
\textbf{Model} & \textbf{Img Quality$\uparrow$} & \textbf{Subject Cons.$\uparrow$} & \textbf{Bg Cons.$\uparrow$} & \textbf{Aesthetic$\uparrow$} & \textbf{Motion Smth.$\uparrow$} & \textbf{Rank Avg.$\downarrow$} \\
\midrule
StableDiffusion~\cite{stablevideo} + CogVideo~\cite{cogvideo} & 75.52 & 78.05 & 85.27 & 59.61 & 97.63 & 4.80 \\
StableDiffusion~\cite{stablevideo} + Wan2.1~\cite{wan} & \textbf{76.93} & 78.54 & 87.43 & 69.75 & 96.67 & 3.40 \\
StoryAdapter~\cite{storyadapter} + CogVideo~\cite{cogvideo} & 76.17 & 72.17 & 88.03 & 63.55 & 98.16 & 4.00 \\
StoryAdapter~\cite{storyadapter} + Wan~\cite{wan} & 75.96 & 75.04 & 88.64 & 73.38 & 97.02 & 3.60 \\
MovieAgent~\cite{movieagent} & 72.09 & 68.61 & 79.84 & 55.40 & 99.01 & 5.80 \\
VoT~\cite{videogenofthought} & 63.85 & 75.11 & 85.78 & \textbf{74.91} & \textbf{99.25} & 3.60 \\
\textbf{\model (Ours)} & 73.64 & \textbf{82.11} & \textbf{88.94} & 64.47 & 98.58 & \textbf{2.80} \\
\bottomrule
\end{tabular}
\end{table*}

\begin{table*}[!htbp]
\centering
\caption{Ablation studies: InfinityStory trained with multi-subject transitions and background injection achieves the best metrics compared to variants without these components.}
\label{tab:vbench_ablation}
\small
\setlength{\tabcolsep}{1pt}
\begin{tabular}{@{}lccccc@{}}
\toprule
\textbf{Model} & \textbf{Img Quality$\uparrow$} & \textbf{Subject Cons.$\uparrow$} & \textbf{Bg Cons.$\uparrow$} & \textbf{Aesthetic$\uparrow$} & \textbf{Motion Smth.$\uparrow$}\\
\midrule
\model & \textbf{73.64} & \textbf{82.11} & \textbf{88.94} & \textbf{64.47} & \textbf{98.58} \\
\quad w/o Background Injection & 72.61 & 78.51 & 87.32 & 62.74 & 98.58 \\
\quad w/o Multi-Character Transition & 72.36 & 81.31 & 88.64 & 63.00 & 97.63 \\
\bottomrule
\end{tabular}
\end{table*}

\subsubsection{Synthetic Data Creation \& Filtering}
\label{sec:dataset}
A hierarchical multi-agent system~\cite{yang2025qwen3} generates structured transition scenarios across five categories: entry, exit, no change, combination, and replacement. We create 10,000 synthetic transition videos targeting underrepresented multi-subject cases. We then generate transition clips from the agentic prompts and filter them using a VLM to curate the final dataset used to finetune our transition model.

\noindent\textbf{Stage 1: Agentic Planning.} A hierarchical LLM system~\cite{minimaxm2,yang2025qwen3} starts by (i) \textit{Scenarios Setup Agent} generates planning documentation for transition mechanics, character interactions, environment planning, and critical generation rules. Shots may begin or end with zero to four characters and conclude with the same or a different number of characters, up to a maximum of four. For example, entry scenarios (0$\to$X) must start with completely empty scenes, location diversity is enforced and all scenarios use anime/cartoon/3D animation styles only. The system outputs 10,000 structured prompts spanning 1-4 characters across five transition categories.
(2) \textit{Video Architect Agent} then generates 40 high-level scenario flavors covering entry, exit, replacement, no-change, and combination cases. (3) \textit{Video Director Agent} generates 250 detailed variations per flavor (10 batches × 25 scenarios), specifying character descriptions, movement choreography, shot types, camera movements, and interaction types; (4) \textit{Video Prompter Agent} synthesizes scenarios into T2V prompts with comprehensive scene descriptions and negative prompts specifying issues to avoid.

\noindent\textbf{Stage 2: T2V Generation.} Wan2.2~\cite{wan} generates 5-second transition videos from the 10,000 prompts generated in stage 1.

\noindent\textbf{Stage 3 VLM Filtering.} A vision-language model Qwen3-VL~\cite{yang2025qwen3} evaluates first and last frames with zero-tolerance character counting: any visible human presence (including partial body parts, distant figures, edge entries) is counted; only shadows, reflections, posters, and statues are excluded. Videos failing the target character counts are discarded. We ended up with 3980 videos ~39.8\% of the original videos. 

\noindent\textbf{Stage 4 FLF2V Training.} We train Wan2.1-FLF2V-14B~\cite{wan} using Low-Rank Adaptation (LoRA) exclusively on the filtered underrepresented multi-character dataset, sampling transition metadata $\tau$ from a balanced prior. This prior forces the dataset to cover rare cases.
\section{Experiments}
\label{sec:exp}

\subsection{Experiment Setup}

\noindent \textbf{Evaluation Dataset.} We use 10 stories from TinyStories~\cite{tinystories} to evaluate our InfinityStory work. These stories are multi-character which occur in different locations. 

\noindent \textbf{Baselines.} We evaluated recent storytelling models such as StoryGen~\cite{storygen}, StoryDiffusion~\cite{storydiffusion}, StoryAdapter~\cite{storyadapter}. The prior baselines fail to create automated movie generation or long form storytelling videos. However,  MovieAgent~\cite{movieagent}, and VideoGen-of-Thought~\cite{videogenofthought} use automated movie generation pipeline but fail to address background consistency and smooth multi-subject consistency.

\noindent \textbf{Evaluation Metrics.} We use VBench metrics~\cite{vbench} as the evaluation metric. More specifically, Image Quality is evaluated using the MUSIQ~\cite{ke2021musiq} image quality predictor; Subject consistency assess
whether character appearance remains consistent by calculating DINO~\cite{caron2021emerging} similarity; Background Consistency evaluates the temporal consistency of the background scenes by
calculating CLIP~\cite{radford2021learning} feature similarity across frames; Aesthetic Quality is evaluated by the LAION aesthetic predictor~\cite{schuhmann2022aesthetic}; Motion Smooth utilizes the motion priors in the video frame interpolation
model~\cite{li2023amt} to evaluate the smoothness of generated motions. The dynamic degree metric is derived from RAFT-based optical flow, which reflects only motion magnitude rather than narrative consistency or semantic correctness. Therefore, we exclude this metric from our main evaluation. Please refer to the supplementary material for a more detailed discussion.

\subsection{Main Results}

Our method achieves the strongest overall performance on VBench, with the \textbf{best average rank 2.80} among all baselines. 
The main advantage comes from two aspects: (1) our scene-preserving video generation strategy, in which the global layout of each scene is grounded through background injection; and (2) accurate character referencing enabled by agent planning and character-aware conditions, with cross-shot consistency further ensured by our multi-character transition mechanism. For example, our model achieves the highest Subject Consistency (82.11) and the highest Background Consistency (88.94), outperforming all other comparison baselines.

Despite the strong consistency, our Image Quality and Aesthetic score are slightly lower than some baselines. 
This is expected for two reasons. 
First, our pipeline operates at 480p resolution, whereas the underlying video generator (Wan2.2) performs best at 720p. 
Second, the editing module (OmniGen2) inherently introduces visual degradation during image-edit operations: while it enables character injection, there would be some visual artifacts in the edited regions, which directly affects perceptual quality scores.

For motion smoothness, our results remain competitive with other baselines, suggesting that the consistency-focused design does not compromise temporal coherence. Our strong performance on consistency-related metrics aligns with this characteristic, while also revealing that VBench may not fully capture higher-level narrative quality or story-driven temporal logic. More results and studies can be found in the supplementary.

\subsection{Ablation Studies}

\noindent\textbf{Effect of Background Injection.} Removing background injection leads to a consistent drop in all metrics, particularly in subject consistency ($\downarrow$3.6) and background consistency ($\downarrow$1.6). This demonstrates that background injection plays a crucial role in maintaining subject and contextual coherence across shots.

\noindent\textbf{Effect of  Multi-Character Transition.} When disabling multi-character transition, the model exhibits reduced motion smoothness and a notable performance drop on aesthetic and motion smoothness. This suggests that transition modeling effectively helps to handle character handovers, allowing smooth transitions while preserving subject integrity.
\section{Conclusions}
\label{sec:conc}
We introduced \model, a framework that enables long-form narrative video generation with world-level consistency and smooth multi-character shot transitions. Through hierarchical multi-agent planning, location-grounded background injection, and a transition model trained on a curated multi-subject dataset, our system directly addresses longstanding challenges of scene drift and abrupt character changes. Empirically, InfinityStory achieves the highest Background Consistency (88.94), highest Subject Consistency (82.11), and best overall VBench average rank (2.80), outperforming all baselines in maintaining stable environments and coherent character dynamics across extended sequences. These results demonstrate that explicit location binding and supervised multi-subject transitions provide a strong foundation for scalable, cinematic video storytelling, paving the way for future systems capable of generating richer and longer narrative experiences.

\noindent\textbf{Limitation.} A remaining limitation is that the FLF2V transition model exhibits limited generalization to unseen character combinations and complex storylines. In the future, we will explore expanding the transition dataset and incorporating multi-prompt supervision to further enhance robustness and narrative flexibility.

{
    \small
    \bibliographystyle{ieeenat_fullname}
    \bibliography{main}
}
\clearpage
\begin{figure*}[t!]
\centering
\includegraphics[width=1\textwidth]{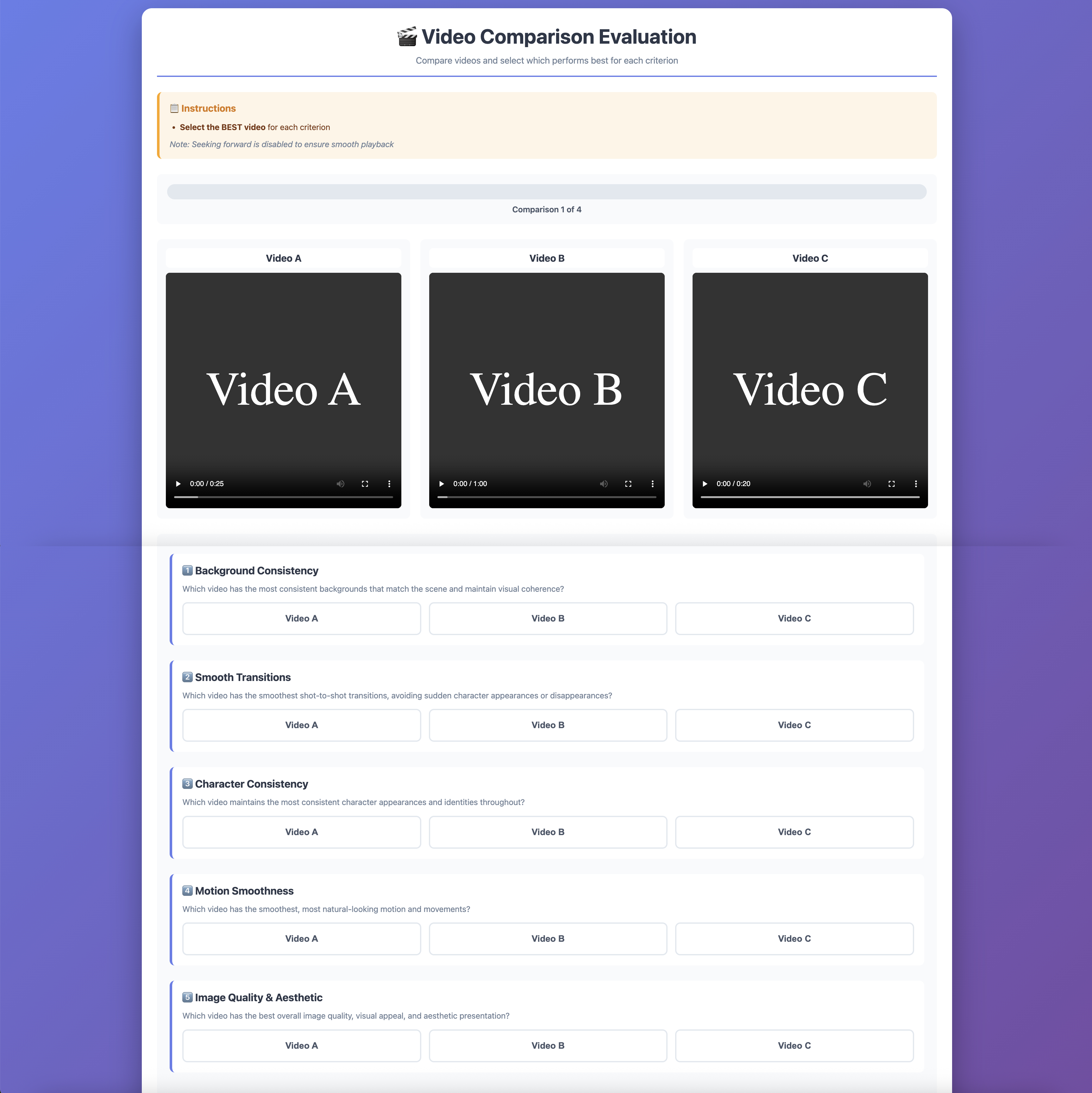}
\caption{Website we developed to collect human evaluations with Video A, B, and C altering between the three methods.
}
\label{fig:human-web}
\end{figure*}
\begin{figure*}[t!]
\centering
\includegraphics[width=0.97\textwidth]{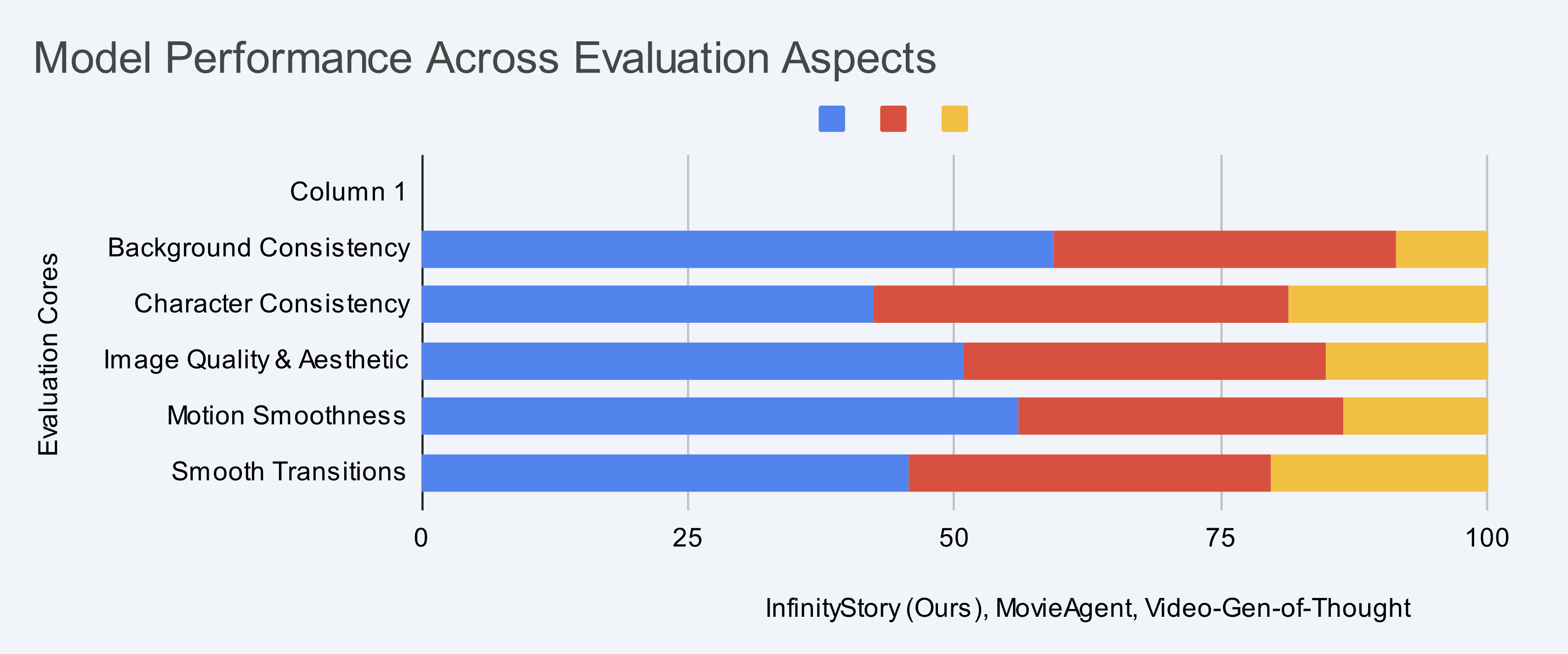}
\caption{Results show that we outperform other methods on human studies.
}
\label{fig:human}
\end{figure*}
\begin{table*}[t!]
\centering
\caption{Human Study Results}
\label{tab:human_study}
\large  
\begin{tabular}{lccc}
\toprule
\textbf{Metric} & \textbf{InfinityStory (Ours)} & \textbf{MovieAgent} & \textbf{Video-Gen-of-Thought} \\
\midrule
Background Consistency     & 59.32 & 32.20 & 8.47 \\
Character Consistency      & 42.37 & 38.98 & 18.64 \\
Image Quality \& Aesthetic & 50.85 & 33.90 & 15.25 \\
Motion Smoothness          & 55.93 & 30.51 & 13.56 \\
Smooth Transitions         & 45.76 & 33.90 & 20.34 \\
\bottomrule
\end{tabular}
\end{table*}

\setcounter{section}{0}
\renewcommand\thesection{S\arabic{section}}
\renewcommand\thesubsection{S\arabic{section}.\arabic{subsection}}
\renewcommand\thesubsubsection{S\arabic{section}.\arabic{subsection}.\arabic{subsubsection}}

\setcounter{figure}{0}
\renewcommand\thefigure{S\arabic{figure}}
\setcounter{table}{0}
\renewcommand\thetable{S\arabic{table}}
\setcounter{equation}{0}
\renewcommand\theequation{S\arabic{equation}}

\section{Experiments}

\subsection{Human Studies.}
We conducted a user study with 20 participants evaluating 20 videos generated by InfinityStory (ours), MovieAgent~\cite{movieagent}, and Video-Gen-of-Thought~\cite{videogenofthought}. The evaluation followed a comparative scheme (Figure~\ref{fig:human-web}), where each participant viewed three videos and selected the best one according to: (1) background consistency and scene coherence, (2) smoothness of shot-to-shot transitions without abrupt character appearance or disappearance, (3) character identity consistency across shots, (4) motion smoothness and naturalness, and (5) overall image quality, visual appeal, and aesthetic presentation.

Figure~\ref{fig:human} and Table~\ref{tab:human_study} show our human study results, where InfinityStory consistently outperforms MovieAgent and Video-Gen-of-Thought across all evaluated metrics. InfinityStory achieves the highest background consistency (59.32 vs. 32.20 and 8.47), motion smoothness (55.93 vs. 30.51 and 13.56), and image quality and aesthetic (50.85 vs. 33.90 and 15.25), indicating stronger scene coherence and visual stability. InfinityStory also leads in character consistency (42.37 vs. 38.98 and 18.64) and smooth transitions (45.76 vs. 33.90 and 20.34), reflecting more natural multi-shot storytelling compared to prior work.

\subsection{Main Results.} 
Dynamic Degree is a VBench metric that measures the magnitude of motion in a generated video using RAFT optical flow~\cite{teed2020raft}. A value of 100 percent corresponds to extremely dynamic scenes such as motorcycle racing or fast car motion. The VBench paper~\cite{vbench} reports that models with high temporal consistency, including strong background consistency, subject consistency, and motion smoothness, often show lower Dynamic Degree because the background remains stable and the motion is controlled.

StoryAdapter combined with Wan~\cite{storyadapter,wan} produces high motion with a Dynamic Degree of 84.73. VoT~\cite{videogenofthought} produces very low motion with a value of 3.50. InfinityStory obtains a moderate value of 53.35. The remaining models fall within similar ranges, with StableDiffusion combined with CogVideo~\cite{stablevideo,cogvideo} scoring 33.58, StableDiffusion combined with Wan2.1~\cite{stablevideo,wan} scoring 60.30, StoryAdapter combined with CogVideo~\cite{storyadapter,cogvideo} scoring 26.71, and MovieAgent~\cite{movieagent} scoring 35.44. This level reflects the design of our method, which focuses on scene continuity and multi shot consistency rather than maximizing motion amplitude. Dynamic Degree should be interpreted together with the temporal consistency metrics rather than as a direct indicator of video quality.

\noindent\textbf{Ablations.} We first experimented with generating keyframes for every shot. In this setting, each transition was generated strictly between two fixed keyframes. This forced the storytelling video to follow the keyframe layout exactly, leaving little room for natural motion or creative interpolation across shots. Our final design keeps keyframes only for odd main shots, while even shots are transition shots generated directly from the last frame of the preceding I2V output. This lets the transition model (FLF2V) start from an actual visual state rather than a predefined keyframe. In earlier ablations, transitions were always two seconds long, which caused characters to rush toward their keyframe positions before the transition ended. This problem motivated the revised structure of using I2V for odd shots and FLF2V for even shots, where transitions evolve smoothly from the final frame of the previous shot instead of being constrained by a fixed keyframe target. 

We initially used Qwen Image Edit 2509~\cite{yang2025qwen3} for character injection. When injecting two or more reference images, we observed noticeable changes in character identity, which is critical for storytelling applications. Replacing it with OmniGen2~\cite{omnigen2} resolved this issue, as the model preserved character identity more reliably during multi-image injection. 

\section{Methods}
\subsection{Storytelling Video Generation.} We used an agentic methodology to generate structured files which was used in the video generation. In this section we highlight the input and output prompt samples.

\begin{figure*}[t!]
\centering
\includegraphics[width=1\textwidth]{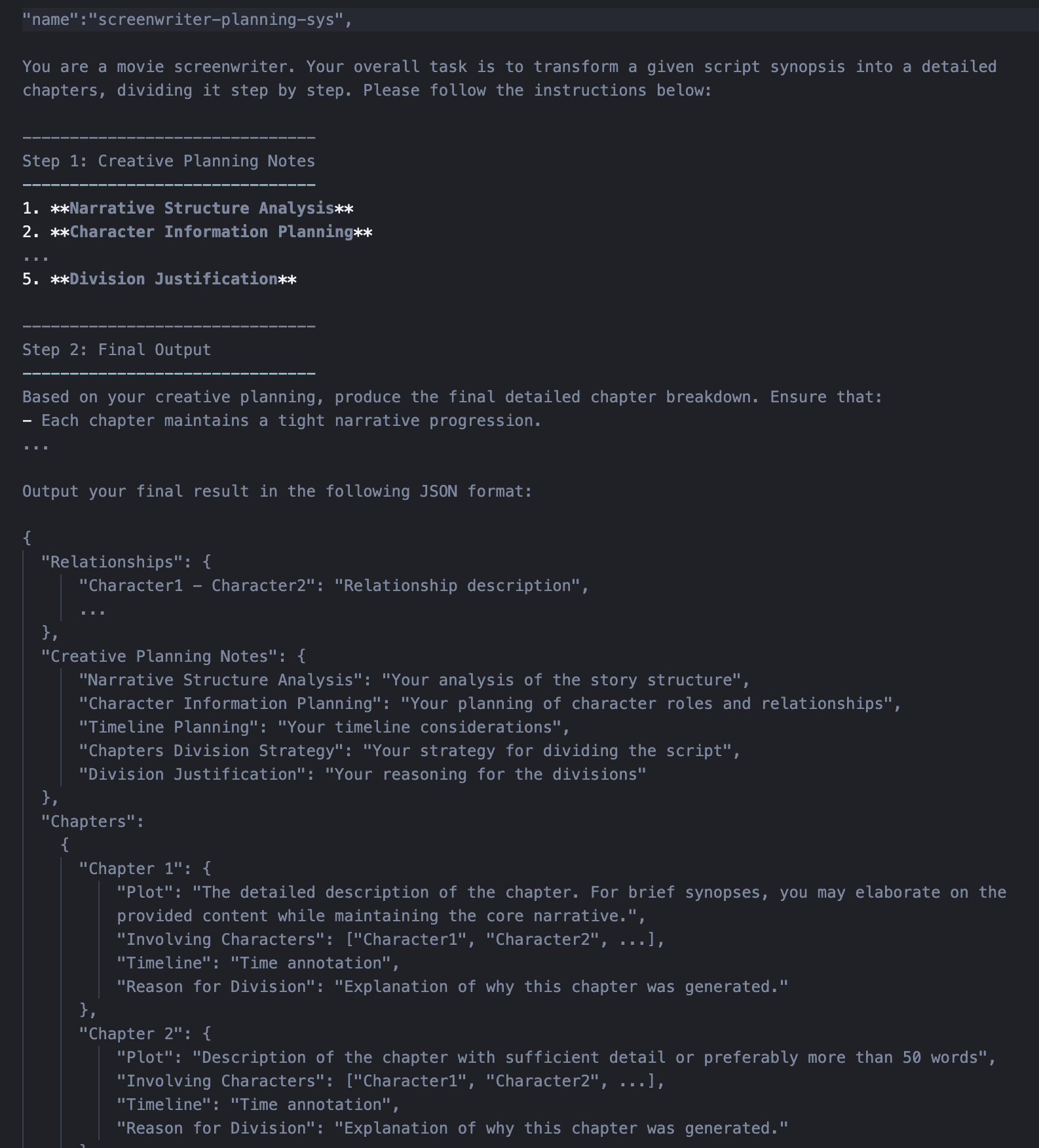}
\caption{Agentic story planning creating the story in the form of chapters
}
\label{fig:prompt-chapter}
\end{figure*}
\begin{figure*}[t!]
\centering
\includegraphics[width=1\textwidth]{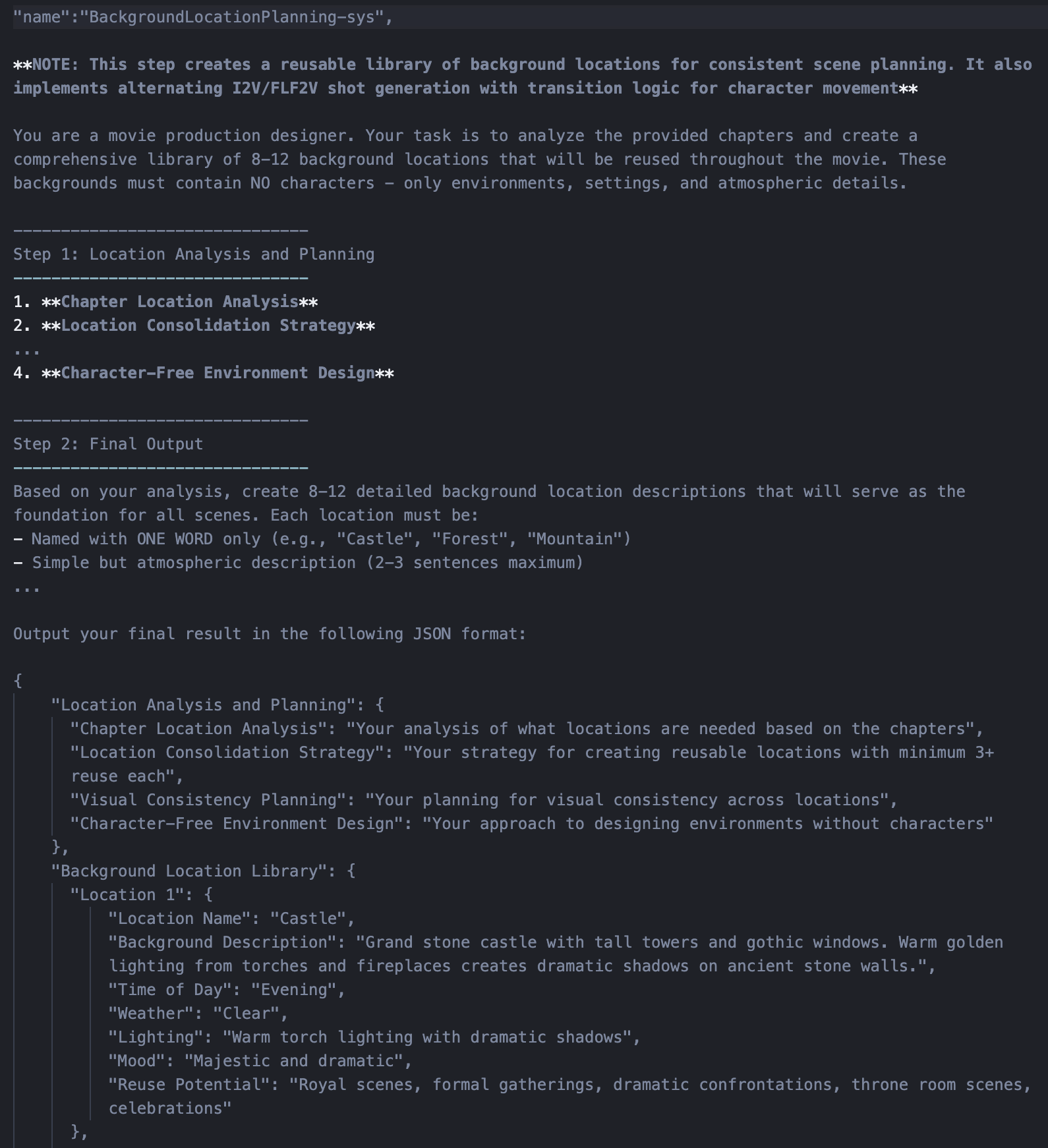}
\caption{Location-based background prompt for background consistency
}
\label{fig:prompt-bg}
\end{figure*}
\begin{figure*}[t!]
\centering
\includegraphics[width=1\textwidth]{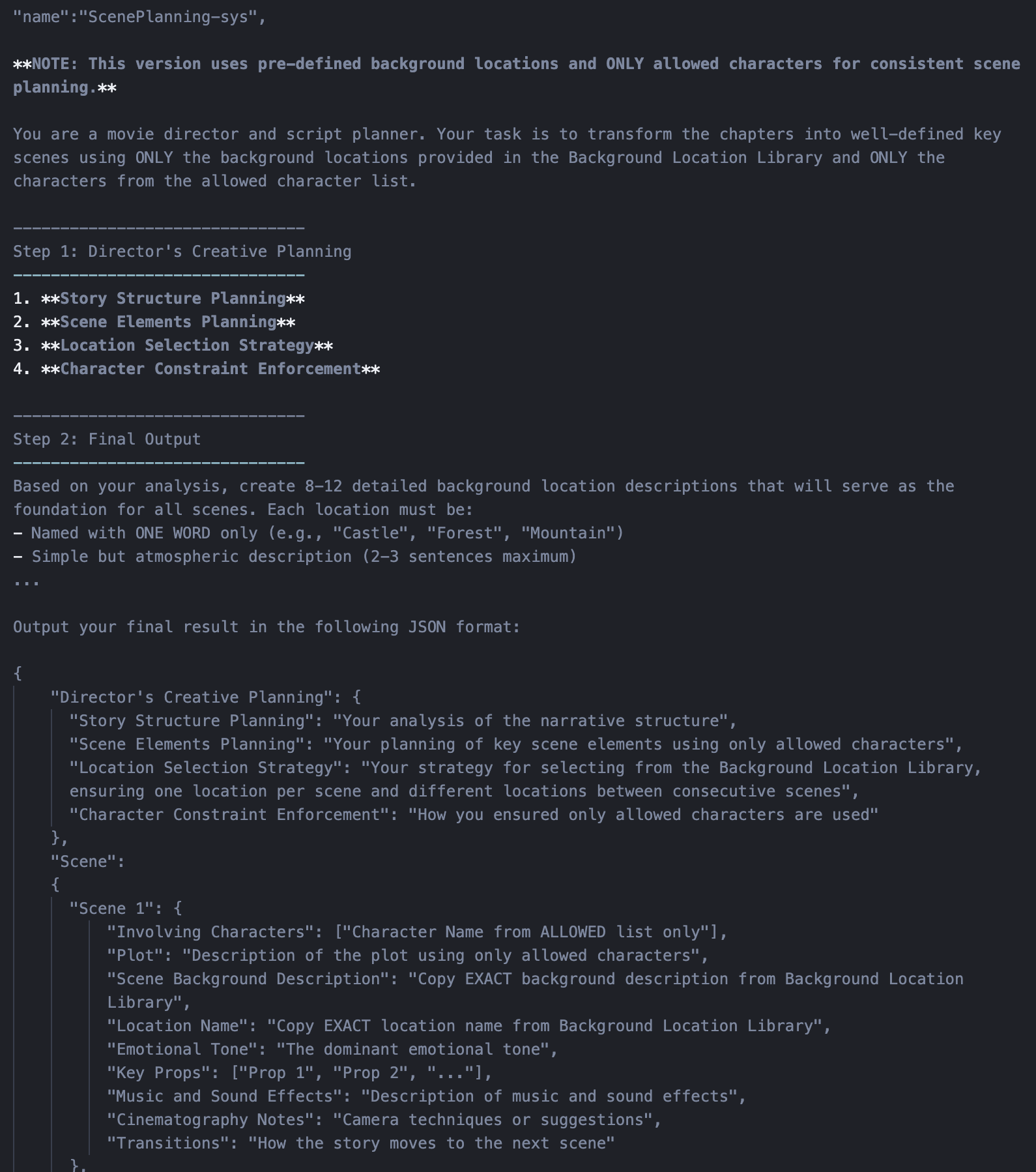}
\caption{Scene planning prompt utilizing the story chapter planning along with the reusable background locations.
}
\label{fig:prompt-scene}
\end{figure*}
\begin{figure*}[t!]
\centering
\includegraphics[width=1\textwidth]{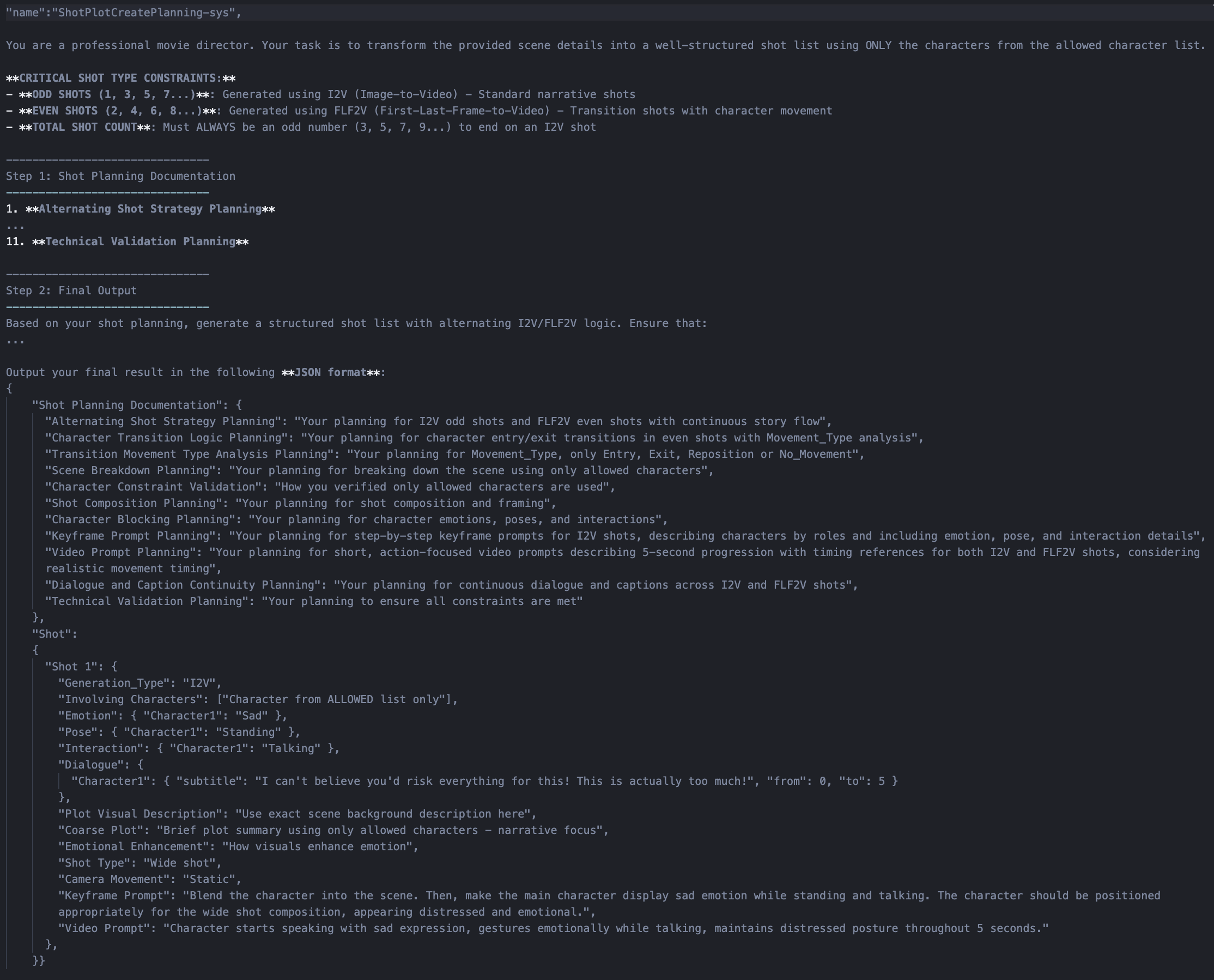}
\caption{Narrative shot planning based on the scenes for the main even shots.
}
\label{fig:prompt-shot-i2v}
\end{figure*}
\begin{figure*}[t!]
\centering
\includegraphics[width=1\textwidth]{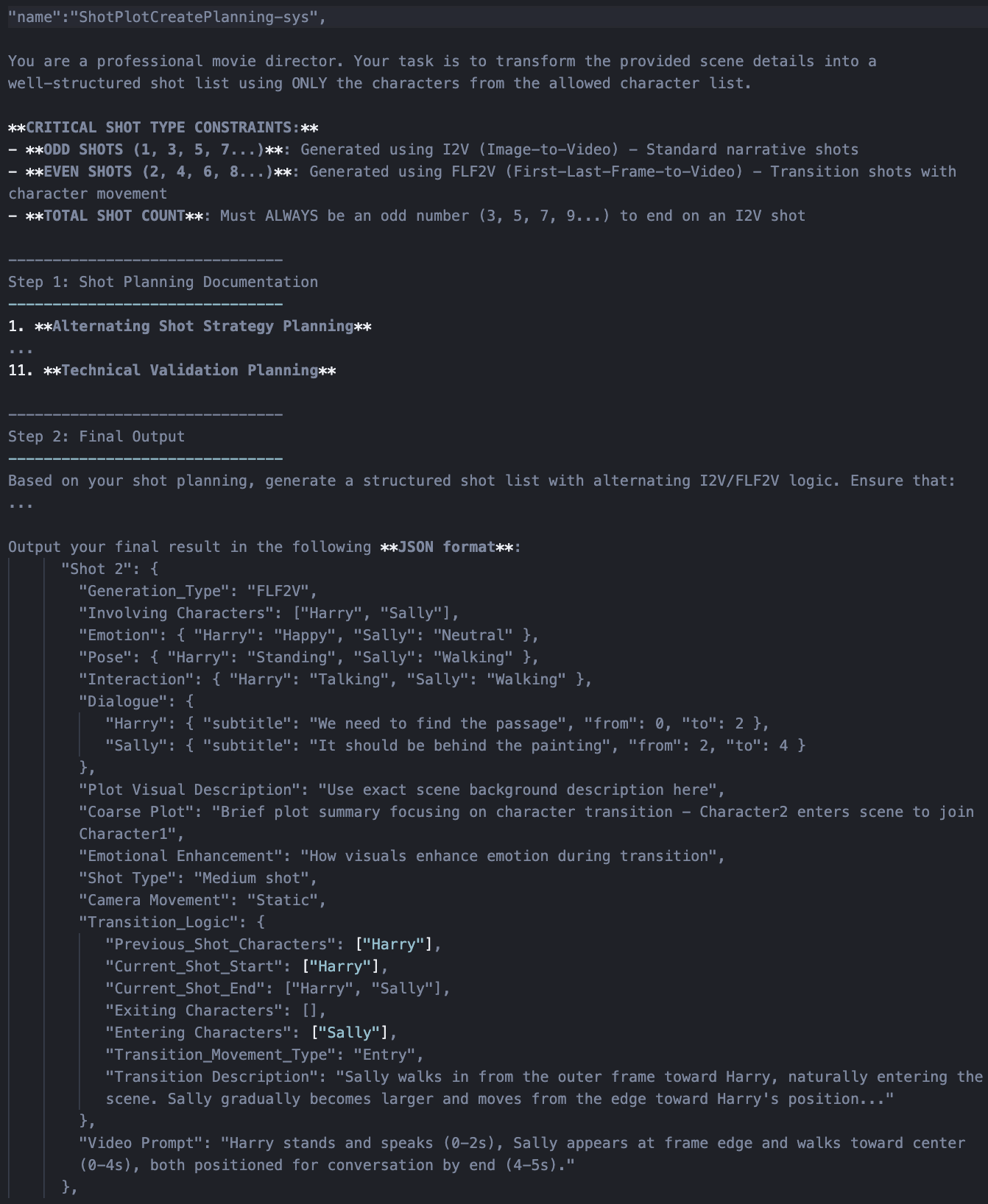}
\caption{Transition shot planning based on the scenes for the odd shots.
}
\label{fig:prompt-shot-flf2v}
\end{figure*}

\noindent\textbf{Agentic Story Planning.} Our storytelling pipeline uses a sequence of system prompts that decompose the user script into structured, machine-readable components. The first stage (“screenwriter planning”) converts the synopsis into chapters using strict JSON constraints. This ensures consistent narrative structure, character tracking, and timeline annotations that later guide scene generation. Figure~\ref{fig:prompt-chapter} provides the system prompt used for the story structure.

\noindent\textbf{Background Location Planning.} The second stage (“background location planning”) creates a reusable library of 8–12 one-word background locations with character-free descriptions. These backgrounds are consistently reused across scenes to maintain world continuity and support background injection during video generation. Figure~\ref{fig:prompt-bg} provides the system prompt used for the location-based background generation.

\noindent\textbf{Scene Planning with Alternating Shot Types.} The third stage (“scene planning”) transforms each chapter into scenes with alternating I2V and FLF2V shots. Odd shots are I2V narrative shots, while even shots are FLF2V transition shots that handle character entry, exit, and repositioning. This structure enforces consistent shot planning, controlled transitions, and location coherence. Figure~\ref{fig:prompt-scene} provides the system prompt used for dividing the chapters in main scenes with consistent background.

\noindent\textbf{Shot-Level Planning.} The final stage (“shot plot planning”) generates detailed shot specifications, including emotions, poses, interactions, camera movement, cinematography notes, dialogue timing, and transition logic. Odd shots contain keyframe instructions for I2V generation, while even shots include explicit transition logic derived from character differences between adjacent shots. This system ensures strict character constraints, stable background reuse, and smooth action continuity across the full video. Figure~\ref{fig:prompt-shot-i2v} and Figure~\ref{fig:prompt-shot-flf2v} provides the system prompt used for the main narrative shots and the transitional shots.

\subsection{Smooth Multi-Subject Shot-to-Shot Transitions}
\begin{figure*}[t!]
\centering
\includegraphics[width=1\textwidth]{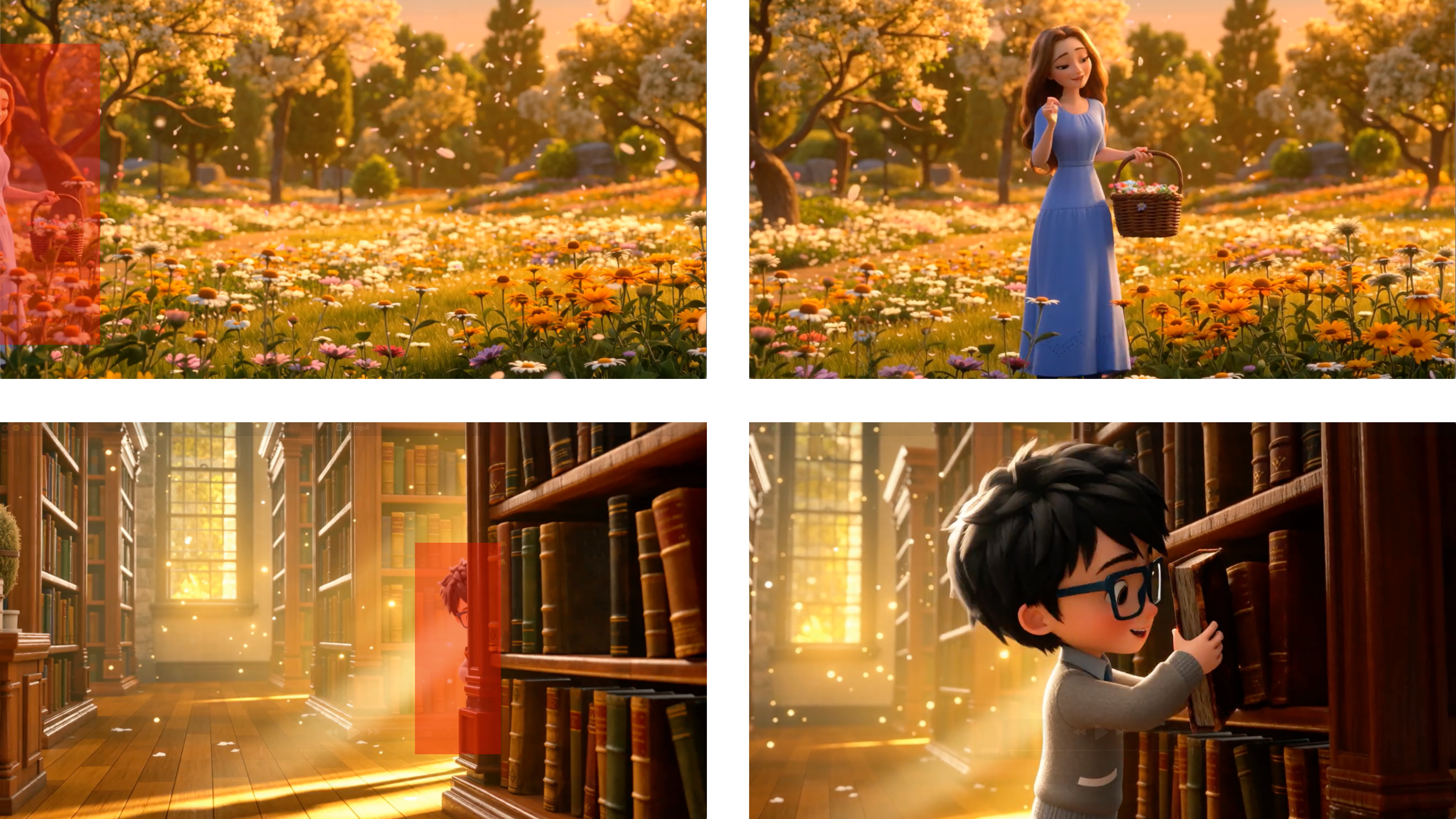}
\caption{Failed transition dataset generation as the character appears partially in the first frame. Character in first frame highlighted in red. Transition from zero characters to one character. These failed videos are filtered by our VLM.
}
\label{fig:transition-failure}
\end{figure*}

\begin{figure*}[t!]
\centering
\includegraphics[width=1\textwidth]{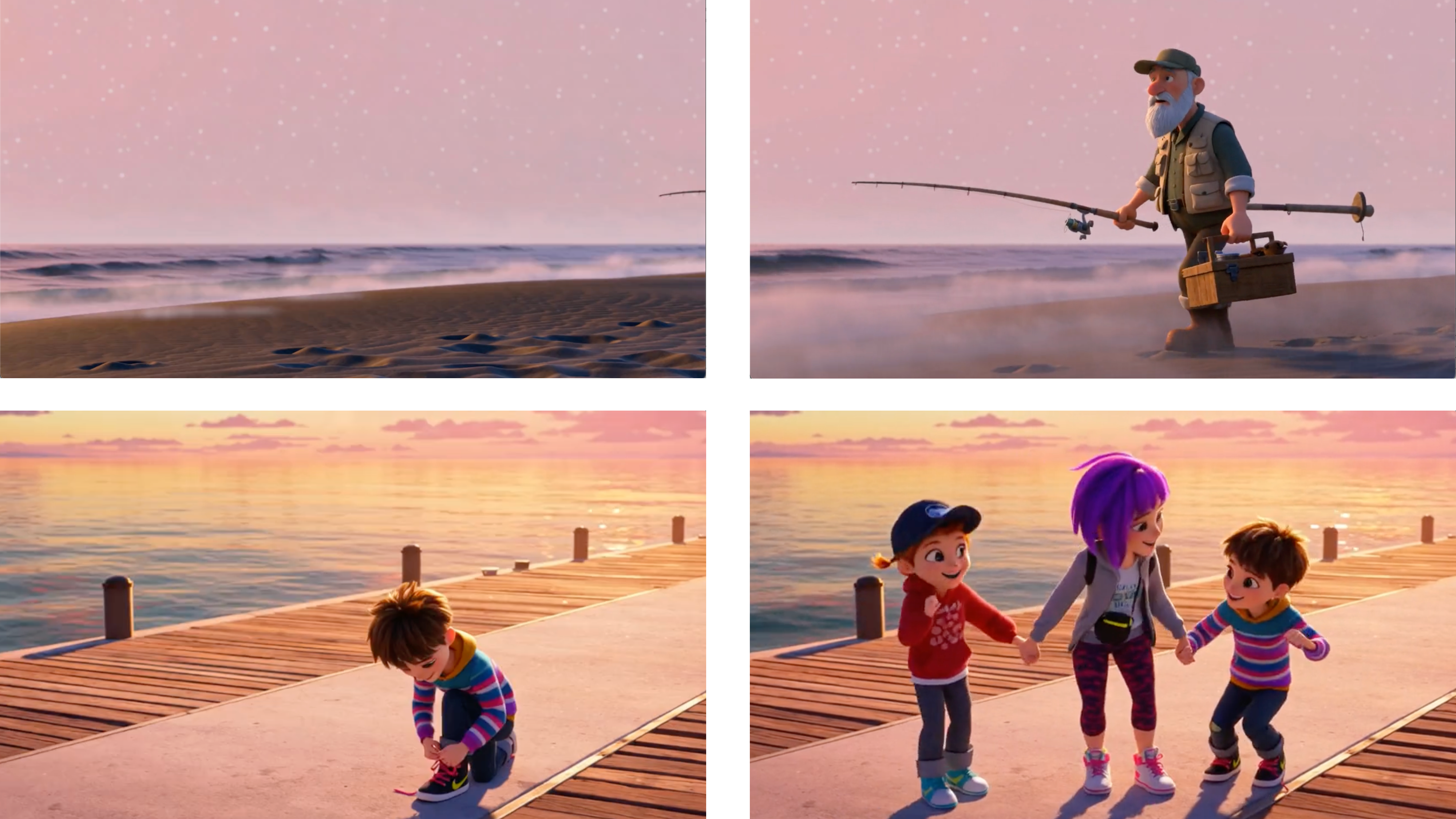}
\caption{Successful transition dataset generation as no character appears in full nor partially in the first frame. First row is a transition from zero characters to one character. Second row is a transition from 1 character to 3 characters.
}
\label{fig:transition-success}
\end{figure*}

We adopt an agentic pipeline that generates structured intermediate representations to guide multi-subject transitions across shots. This enables the model to reason about character entry, exit, and repositioning while preserving scene continuity. Figure~\ref{fig:transition-failure} shows a failure transition scenario automatically filtered by our VLM, while Figure~\ref{fig:transition-success} illustrates a successful transition used to train the FLF2V module.

\paragraph{Multi-Subject Transition Dataset.}
In the main paper, all experiments were conducted using a dataset of 10{,}000 multi-subject transition sequences. To the best of our knowledge, this is the first dataset specifically focused on structured multi-subject shot-to-shot transitions for long-form video generation. The full 10{,}000-sequence dataset will be uploaded and released publicly to support future research.


\end{document}